\documentclass{article}

\usepackage{PRIMEarxiv}

\usepackage[utf8]{inputenc} 
\usepackage[T1]{fontenc}    
\usepackage{hyperref}       
\usepackage{url}            
\usepackage{booktabs}       
\usepackage{amsfonts}       
\usepackage{nicefrac}       
\usepackage{microtype}      
\usepackage{lipsum}
\usepackage{fancyhdr}       
\usepackage{graphicx}       
\graphicspath{{media/}}     

\title{﻿Development of Image Collection Method Using YOLO and Siamese Network}

\author{
  ﻿Chan Young Shin1, ﻿Ah hyun Lee2, ﻿Jun Young Lee3 , ﻿Ji Min Lee4, ﻿Soo Jin Park5 \\
  ﻿Department of Artificial Intelligence and Data Sciencen \\
  ﻿Sejong University\\
  ﻿Seoul, Republic of Korea\\
  \texttt{\{﻿Chan Young Shin1, ﻿Ah hyun Lee2\}﻿23012094@sju.ac.kr,﻿ahyunclara@sju.ac.kr} \\
   \And
  ﻿
}

\begin{document}
\maketitle

\begin{abstract}
As we enter the era of big data, collecting high-quality data is very important. However, collecting data by humans is not only very time-consuming but also expensive. Therefore, many scientists have devised various methods to collect data using computers. Among them, there is a method called web crawling, but the authors found that the crawling method has a problem in that unintended data is collected along with the user. The authors found that this can be filtered using the object recognition model YOLOv10. However, there are cases where data that is not properly filtered remains. Here, image reclassification was performed by additionally utilizing the distance output from the Siamese network, and higher performance was recorded than other classification models. (average \_f1 score YOLO+MobileNet 0.678->YOLO+SiameseNet 0.772)) The user can specify a distance threshold to adjust the balance between data deficiency and noise-robustness. The authors also found that the Siamese network can achieve higher performance with fewer resources because the cropped images are used for object recognition when processing images in the Siamese network. (Class 20 mean-based f1 score, non-crop+Siamese(MobileNetV3-Small) 80.94 -> crop 
 preprocessing+Siamese(MobileNetV3-Small) 82.31) In this way, the image retrieval system that utilizes two consecutive models to reduce errors can save users' time and effort, and build better quality data faster and with fewer resources than before. 
\end{abstract}


\section{I.INTRODUCTION}
In the era of big data, refined image collection for AI learning is necessary. There are Google Data Search, Kaggle, Awesome Public Datasets Github, Data and Story Library, etc., and the formats are txt, csv, jpg, API calls, etc.

The disadvantage of web-crawled data is that there is a high possibility that there is inherent noise data in the image that does not match the label. This can be a factor that causes performance degradation when using web-crawled data for learning a vision AI model.

In Fig. 1.(a), the Training Accuracy is similar when comparing ‘Noisy w/o. Reg.’ and ‘Noisy w. Reg.’. However, the Test Accuracy is clearly higher in ‘Noisy w/o. Reg.’ than in ‘Noisy w. Reg.’. In (b), the classification model initially correctly classified the panda image as panda, but when an adversarial attack that adds noise is performed, it is classified into a different class with high confidence.

    \includegraphics[width=0.45\linewidth]{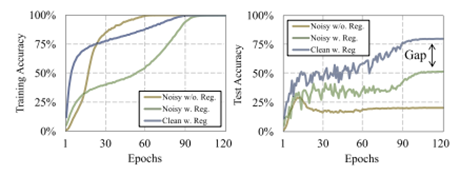}       (a)
            \includegraphics[width=0.45\linewidth]{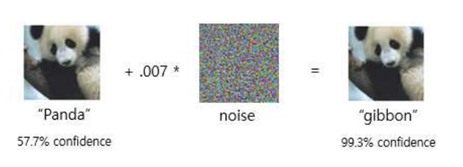}(b)

\textbf{Fig
.1. (a) Graph showing training and test accuracy of WideResNet-16-8 trained on the CIFAR-100 dataset [1] (b) Example of changing the results of a classification model by adding noise to an image [2]}

Web-crawled data contains fine-grained images that are difficult to distinguish between similar classes, especially unlike base-class images.

Xiaoyu Wange [3] considered complex image backgrounds as one of the main factors contributing to variation within the same class in the classification problem of fine-grained images such as large-scale automobile web-crawled data. Accordingly, he proposed the object-centric sampling (OCS) method, which removes the image background by sampling the image window based on object location information. This improved the classification accuracy by allowing focus on the specific features of objects existing in fine-grained images. In this way, when applying a filtering strategy such as image refinement to web-crawled data, noisy data can be effectively selected to help learning.

The advantage of the filtering strategy is that noisy data can be effectively removed, but this can also lead to the data deficiency problem of removing learnable images. When the alignment level is expressed as the consistency between the image and the content, if the level is high, only images with effectively removed noise remain, but learnable images may also be removed depending on the user's task. On the contrary, if the level is low, many learnable images remain, but there is a high probability that images with noise will also be present. In the image captioning area, to solve this problem, a captioner that can control the alignment level was developed by learning image-text pairs with different alignment levels among web-crawled data. [4] This controllable alignment-level model allows the user to specify the noise-robust level according to the user's task. However, this requires a process of labeling and re-training the alignment level every time new data is crawled.

Additional training reduces the real-time performance of the framework and increases the amount of computation and calculation time.

Therefore, we proposed the Siamese Network [5] as a method that can solve this problem with one-shot learning without having to label and re-train new crawled data. Siamese Network can calculate the distance between a given web-crawled image and an anchor image using one-shot learning. Siamese Network is related to similarity learning, which evaluates how similar two objects are. Similarity learning does not learn to label inputs and classify them but evaluates the similarity between two inputs. When evaluating similarity, the Euclidean distance between feature vectors in the representation space is used. Therefore, one-shot learning is a method that evaluates the similarity between a new image and an anchor image using the features of the network itself without having to re-train a pair of new images and their corresponding labels.

    \includegraphics[width=0.5\linewidth]{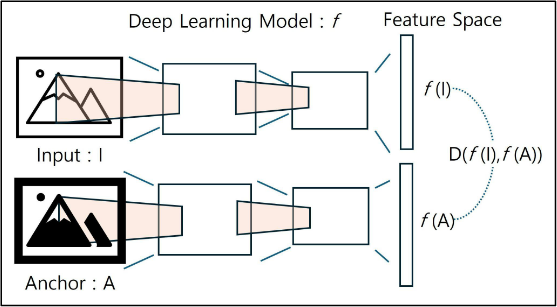}

\textbf{    ﻿Fig.2. Siamese Network architecture : input image I and anchor image A are each input to deep learning model which share weights and return to feature vectors f(I), f(A) in feature/representation space, output becomes distance of these vectors D(f(I),f(A))}

The characteristic of Siamese network is that it can be usefully used in image retrieval applications. Even if the number of new images is small, the classification task can be performed by comparing the input distance and the output distance with the user-specified distance threshold.

Here, the balance between the amount of data provided and the degree of noise removal can be adjusted according to the value of the user-specified distance threshold.

In Fig. 3, when the distance threshold is low, only images with a lower distance than this are classified as keywords. Therefore, only images with a high alignment level between the keyword and the image content are included as keywords, so the number of images classified as keywords is small. On the contrary, when the distance threshold is high, all images with a lower distance than this are included, so the number of images classified as keywords is large.

In other words, the lower the distance threshold, the fewer images that can be used as keywords, while the higher the threshold, the less strictly noise-robust images, but the more images that can be used as keywords.

Therefore, users should carefully consider the trade-offs between data deficiency and noise-robustness when specifying the distance threshold.

     \includegraphics[width=1\linewidth]{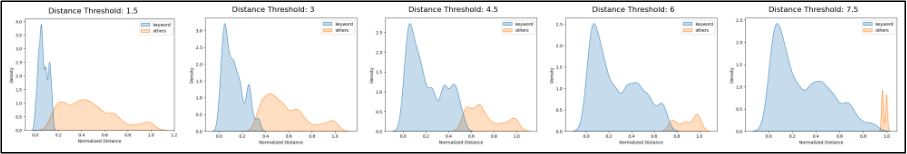}

\textbf{Fig. 3. 100 web-crawled images of aeroplane class classified by distance threshold 1.5 to 7.5 of Siamese Network’s distance with given anchor image : The blue area shows web-crawled images classified as keywords; the orange image shows images classified as non-keywords. The distance threshold is inversely proportional to the strictness of the comparison of the Siamese network. A classifier in the web framework selects images as keywords whose distance is lower than the threshold. }

\textbf{(1) Framework Proposal}

The image data construction framework we developed is a process of selecting appropriate images by cropping images searched by users with keywords using the YOLO model and comparing them with local anchor images through the MobileNet-based Siamese network.

We constructed the framework to solve two problems that can occur in image collection through web crawling: the fine-grained classification problem and the inherent noise problem.

1) To solve the problem of difficulty in distinguishing between similar classes in fine-grained classification, object-centric classification using the object recognition model Yolo was performed, and after this, 2) to solve the inherent noise problem of the classification result and image content, the distance is output through comparison with the keyword image in the Siamese network, and the reclassification process is used as a major component of the framework to remove those below a certain standard.

\textbf{(2) Experimental performance improvement}

Siamese Network showed better classification performance than simply using YOLO or YOLO+MobileNet, and Siamese Network achieved sufficient performance with a lightweight model with fewer resources by using images produced through YOLO preprocessing as input.

We applied YOLO's crop preprocessing and Siamese Network's image matching to our developed web crawling image retrieval application. In this process, we emphasized that a lightweight Siamese network can be constructed through crop preprocessing by applying a lightweight MobileNet-v3-small model to Siamese network through YOLO's crop processing, and through this, users can perform whole image matching for various image objects with a lightweight model in various environments, such as AI mobile.
      \includegraphics[width=0.95\linewidth]{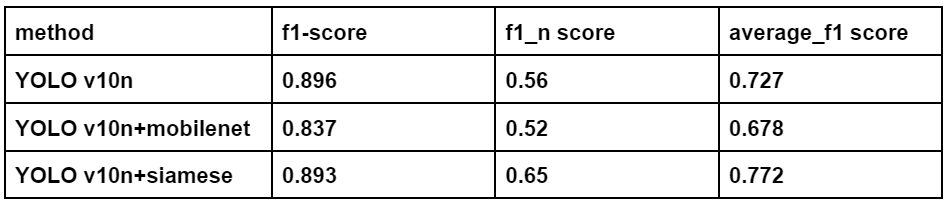}(a)
\includegraphics[width=0.95\linewidth]{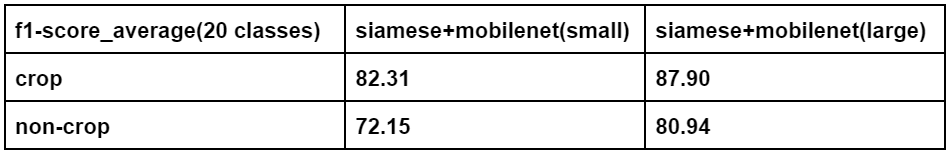}(b)
\textbf{Table.1. performance evaluation of (a) YOLO, YOLO+Mobilenet, YOLO+Siamese network (b) kind of backbone model of Siamese network(MobileNetV3-large, small) and whether image was cropped or not.}

 When a user inputs a keyword, our framework uses an object detection model to crop the recognized object part and then passes it through a Siamese network to determine whether it matches the keyword. Here, the keyword can be a condition. For example, when a user searches for ‘images of people and sofas together’ as a keyword, the image crawler crawls the corresponding images and passes them to the object detection model, which then detects objects (e.g., people, sofas), classifies them into classes, and passes them to the DB. In the DB, the Siamese network model trained on the keyword image secondarily classifies whether the image is a correct keyword image. After that, the final images are returned to the user through the DB. After the above process is completed, objects that match the conditions provided by the user can be selected and provided through the DB. In other words, users can receive photos with various conditions according to their purposes. 

\section{\textbf{II. RELATED WORK}
}
\textbf{A. Image matching}

Previously, the Siamese network was used for facial image verification, which matches a given image with a local image. Yatharth V. Kale et al.[6] used it for facial recognition using object detection. Here, a cropped human image with the YOLO model is passed through the Inception model, which is trained with human images in the internal database, to determine who it is. The Siamese network structure is used to compare each feature of the given cropped image and the database image with the Inception model. Through this, the similarity between the user and several people in the database is calculated, and the person with the lowest similarity is output.

Ivan Bakhshayeshi et al.[7] used the image matching of the Siamese network for deep learning facial recognition for cattle reidentification. To complement the various problems (vulnerable to loss, failure, and misidentification or improper substitution) that exist in radio frequency identification (RFID) ear tags used for livestock recognition and management, we proposed a cattle recognition system that combines the You Only Look Once version 5 (YOLOv5) algorithm for cattle face detection and the Siamese neural network (SNN) for subsequent recognition.

In this case, the part where the image given by the YOLO model is cropped and the process of selecting it through the Siamese network based on the classification model are similar to our framework, but we did not limit the application of the Siamese network model to a specific object but applied it to various image objects. In addition, we developed an image retrieval application that can control the strictness of noise-robustness by allowing the user to set the distance threshold that determines how far the distance from the Siamese network will be judged as a keyword.

\textbf{B. Image Retrieval Application}

After that, there were attempts to use the Siamese network for real-time object tracking, image retrieval, and whole image matching. Iaroslav Melekhov et al. [8] compared the existing image classification baseline models with sHybridNet that applied Hybrid CNN to the Siamese network and showed performance improvement, and presented a method of applying the Siamese network to generic image matching. They suggested that an image matching method using a Siamese network like this is likely to be used in image retrieval applications.

There have been many cases of data retrieval applications using web crawling[9]. There are cases of using web crawling for web archiving[10], detecting near-duplicates[11], or  weakly supervised learning. Furthermore, they built an imageDB using web crawling.

Hwang et al. [12] developed a system that updates the system by transfer learning with appropriate train data created by putting the refined image received through the image crawler through an object detector (image annotator) using YOLO and the label set that classified it into a trainer server.

Jeongbin Hwang et al.[13] collected images related to construction monitoring, auto-labeled them using a semantic segmentation model, and then put them into imageDB for learning.

While these studies created an imageDB server that automatically trains crawled images using a deep learning model for self-labeling, we focused on creating a framework that builds image data optimized for AI learning by users without separate labeling.

Xiaoyu Wang[3] considered complex image backgrounds as one of the major factors contributing to intra-class variation and proposed an object-centered sampling (OCS) method that samples image windows based on object location information. It improved the top-1 accuracy to 89.3\% (up from 81.6\%) in a large-scale detailed automobile classification dataset.

Inspired by this, we extracted objects using an object recognition model for fine-grained web-crawled data classification and performed classification based on their features. The study performed detection with regionlets using R-CNN with scale-invariant property for saliency-aware object detection.

On the other hand, we used YOLO, a one-stage model, for real-time performance of the web framework and stored images cropped by YOLO in the DB so that object images could be delivered to users.

In addition, we showed that YOLO's crop preprocessing before inserting it into Siamese Network showed sufficient performance even when using a lightweight model as the backbone model of Siamese Network.

\section{\textbf{\textbf{III. METHOD} }
}
\textbf{﻿Web framework }

    \includegraphics[width=1\linewidth]{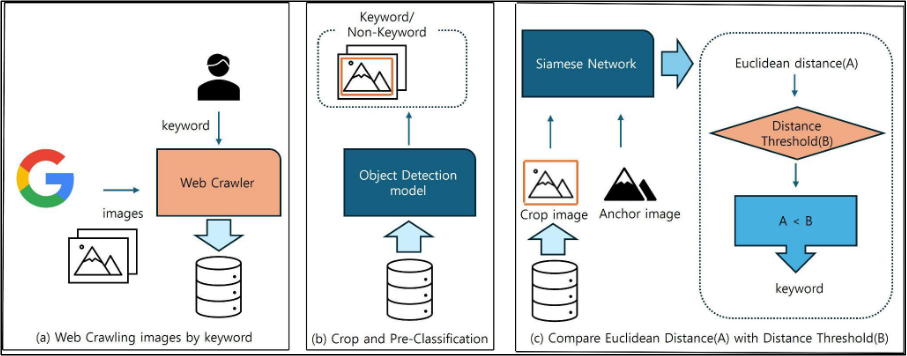}

 Fig. 4. Python Flask-based image data construction web framework (a) Inputs image data keywords to be searched and constructed from users, and based on these, a web crawler based on Python Selenium and Beautiful Soup stores images (I) crawled from Google in the DB. (b) In the DB, the recognized object parts are cropped and classified and re-saved after passing through the object detection model, Yolo. Through this process, images with only the parts corresponding to the keyword cropped are provided to the user, or images with specific objects are returned according to various conditions set by the user. (c) The pre-classified cropped image in the DB and the anchor image of the keyword are simultaneously input into the Siamese Network, and the distance (A) between the feature vectors generated through the backbone network, MobileNetV3, is calculated and compared with the distance threshold (B). If A < B, it is classified as the final keyword. 

 The object detection model uses YOLOv10[14]. YOLO is a 1-stage model that can detect objects in real time by performing classification and localization simultaneously. The dataset used to train the model is the PASCAL VOC dataset 2012 and 2007[15]. The dataset contains 27,652 images (training: 16,551; validation: 4,952) of annotations in 20 classes. The dataset was trained with the following parameters: batch size 64, epochs of 100, and weights of yolov10.pt (pre-trained weight).

    \includegraphics[width=1\linewidth]{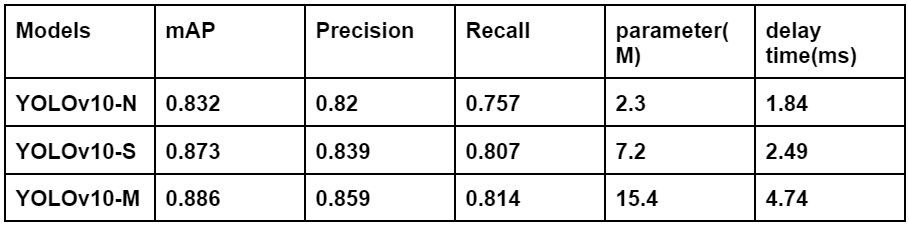}

\textbf{Table.2. Performance evaluation of YOLOv10}

We trained various models according to the parameter size of YOLOv10 on the VOC dataset. We found that the mAP of YOLOv10-N, which has a parameter size that is about 6.7 times smaller than YOLOv10-M, is about 0.054 lower than YOLOv10-M, but the delay time is reduced by 2.9 (ms). In addition, YOLOv10-N was used because the web framework was created for users to easily access the web from their own resource environment.

\textbf{Siamese Network}

The purpose of this framework is to reclassify images classified by YOLO through Siamese network[10]. Therefore, some of the cropped images from the existing YOLO object recognition are used for training. The total number of datasets used for training Siamese Network is 26,360 (train: 20,709, validation: 5651). The number of anchor images used in each train and validation is 1, and the rest are others images. At this time, the number of support sets (classes) used for others is 20, including keyword and non-keyword labels. If it is a keyword label, the Euclidean distance with the anchor image is labeled as 0, and if not, as 1.

In the inference process in Fig. 1.(b), (2) after YOLO classifies it as a keyword or not, (3) the query image and anchor image are input together to Siamese Network, and the similarity between the two images is calculated as a distance.

At this time, if the query image belongs to the space classified by YOLO as a keyword class, if the distance is greater than the threshold distance, it is determined that it is not actually a keyword class image and is removed from the keyword class space. Or, if the query image belongs to the space classified by YOLO as a non-keyword class and the distance is less than the threshold distance, it is determined that it is actually a keyword class image and is moved to the keyword class space.

Siamese Network passes two input data through the backbone network to obtain an embedding vector for each input, and learns the difference by calculating the distance between the two embedding vectors. If the two inputs are in the same class, the distance is adjusted to be close, and if they are in different classes, the distance is adjusted to be farther using contrastive loss. At this time, Euclidean distance is used as a similarity, symbolizing a positive pair. It is not easy to obtain the Euclidean distance between high-dimensional data and causes the curse of dimensionality problem.

Therefore, a deep neural network is used as the backbone network to reduce high-dimensional data to low dimensions. A deep neural network has a powerful dimension reduction function that can embed high-dimensional data such as images into low-dimensional features. [16]

Siamese Network embeds each pair of input images (anchor, others) into a low-dimensional feature using a deep neural network and then learns using contrastive loss as a cost function that makes the Euclidean distance between positive pairs close and negative pairs far in the feature representation space. The idea is to have a deep neural network learn the optimal feature representation that makes positives as similar as possible and negatives as far away as possible.

\[\frac{1}{2}(y\bullet D(I,A)^{2}+(1-y)\bullet(m-D(I,A))^{2})\]
\textbf{Fig.5. loss function} \textbf{Given a pair of an image I  and anchor image A as input, D(I,A) be the Euclidean distance between their feature vectors as output of Siamese Network[5]}

y is the label (1 if the pair is similar, 0 if dissimilar). The range of y is from 0 to 1.

\textbf{ loss function}

the loss function should still ensure that similar pairs have a small Euclidean distance and dissimilar pairs have a large Euclidean distance.

For similar pairs (y=1): The term $y\bullet D(I,A)^{2}$ ensures that the loss is minimized when the Euclidean distance $D(I,A)$ is small.
For dissimilar pairs (y=0): The term $(1-y)\bullet(m-D(I,A))$ (with m being a large constant) encourages larger distances, ensuring that dissimilar pairs are pushed apart.

In the final inference stage, reclassification is performed by comparing distance(A) and threshold(B) based on the results previously classified by YOLO. To summarize:

if YOLO result is keyword:A>B then classify as non-keyword;

if YOLO result is non-keyword:A<B then classify as keyword.

In the case of pre-classified keyword by YOLO:

When distance A is greater than threshold B, reclassify as non-keyword**.

That is, if A>B, reclassify as non-keyword.

In the case of pre-classified non-keyword by YOLO:

When distance A is less than threshold B, reclassify as keyword**.

That is, if A<B, reclassify as keyword.

Criteria for selecting distance threshold

Here, depending on the value of the distance threshold, m, the balance between the amount of data provided to the user and the degree of noise removal can be adjusted, which is related to the criteria for selecting the threshold.

We input the val dataset into the model and set the threshold where FP becomes 0 among the output distances to the lowest distance and the threshold where FN becomes 0 to the highest threshold. At realism level 5, even if it is a photo similar to the keyword (ex. a model of the keyword), it should not be returned to the user, and the object that is the real keyword should be returned. Therefore, the point of the realism level 5 should filter out images that are as similar to the keyword as possible, so the threshold was set to the distance where FP=0, PPV(TP/(TP+FP)), and the ratio of the correct images among those that the model said were correct starts to almost 1. Similarly, at realism level 1, the model should return models that are as similar to an airplane as possible to the user but should include images that are real keywords. Therefore, in order for the model to return all images similar to the keyword, including the keyword images, FN=0 should be used for sensitivity (TP/(TP+FN)), and the point where the ratio of all images that are real keywords classified as keywords becomes 1 was set to the realism level 1.

    \includegraphics[width=0.4\linewidth]{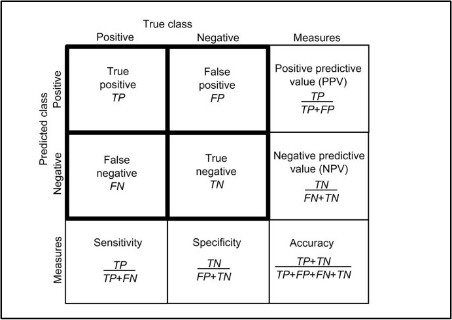} 
    \includegraphics[width=0.4\linewidth]{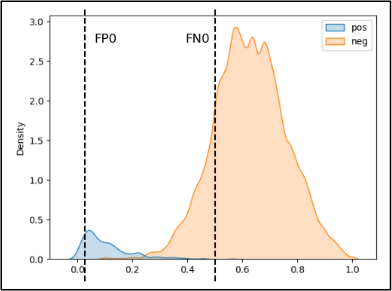}
    
     (a)   \; \;\;\;\;\;\;\;\;\;\;\;\;\;\;\;\;\;\;\;\;\;\;\;\;\;\;\;\;\;\;\;\;\;\;\;\;\;\;\;\;\;\;\;\;\;\;\;\;\;\;\;\;\;\;\;\;\;\;\;\;  (b)
   
    \textbf{Fig.6. (a)confusion matrix and performance metrics (b) Density Graph of Siamese Network Results, x axis means distance output and y axis means density of sample  blue area means true labels and orange area means false labels FP0 line means distance threshold when FP=0 FN0 line means distance threshold when FN=0}

 Distance threshold based on the vertical line, the right side with a greater distance than this is determined as negative, and the left side with a smaller distance is determined as positive. Based on the baseline when FP=0, the left area is not included in the positive areas, so the PPV is 1. Similarly, based on the baseline when FN=0, the right area is determined as negative and the left is determined as positive. Based on this baseline, the right side is not included in the positive areas (pos). In other words, since FN is 0, the sensitivity is 1. 
 
    In Fig. 3, when the distance threshold is low, only images with a lower distance than this are classified as keywords. Therefore, only images with a high alignment level between the keyword and the image content are included as keywords, so the number of images classified as keywords is small. Conversely, when the distance threshold is high, all images with a lower distance than this are included, so the number of images classified as keywords is large.

In other words, the lower the distance threshold, the fewer images that can be used as keywords, while the higher the threshold, the fewer images that can be used as keywords but are less strictly noise-robust.

Therefore, users should carefully consider the trade-offs between data deficiency and noise-robustness when specifying the distance threshold. Low distance threshold (e.g., 1.5): Only images with very high similarity are classified as keywords, so only a small number of images that are very similar to the anchor image are selected as keywords. In this case, images are strictly selected, so noise-resistant classification is achieved, but the number of images selected is small. This corresponds to **high realism level (e.g., Reality Level 5)**. That is, only images that are almost exactly the same as the actual keyword are selected.

High distance threshold (e.g., 7.5): Since even images with low similarity are classified as keywords, many images are selected as keywords, but they are also likely to contain noise. This corresponds to **low realism levels (e.g., Reality Level 1)**, which classifies images less strictly, so more images are included, but images that do not match the anchor image may also be included.

\textbf{Description by Reality Intensity Level:}

    \includegraphics[width=0.7\linewidth]{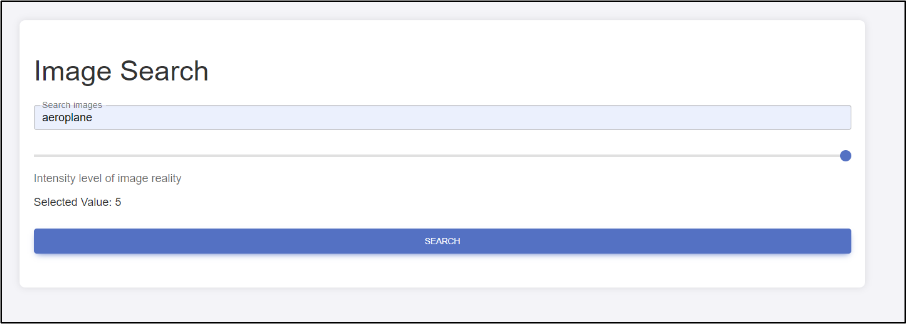}

    \includegraphics[width=0.7\linewidth]{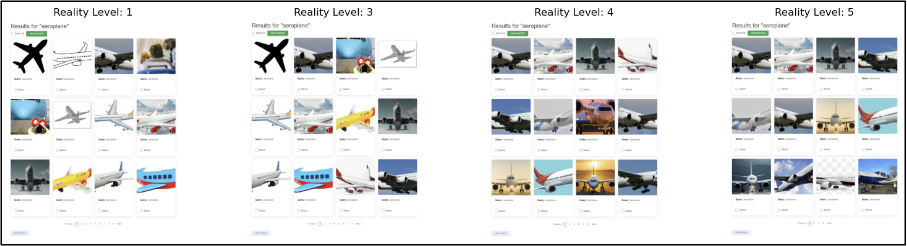}

\textbf{Fig.7. Reality Level 1: Uses the highest distance threshold (7.5). At this level, many images are classified as keywords, but some of them are noisy and less similar to the anchor image. Reality Level 3: Uses a medium threshold (4.5) to classify a moderate amount of images as keywords. It includes both highly similar and somewhat less similar images, balancing noise and accuracy. Reality Level 5: Uses the lowest distance threshold (1.5), so only a few images that are very similar to the anchor image are selected as keywords. This provides the most noise-resistant results but fewer classified images.}

In conclusion, **low threshold (high realism level)** has less noise and higher precision, but fewer images are selected, while **high threshold (low realism level)** has more noise but includes more images. Therefore, users should choose a threshold considering the tradeoff between data scarcity and robustness to noise.

\textbf{B. Siamese Network Architecture}

The Siamese Network used has 20 support sets (classes) to classify, and the learning image uses some of the images output after object recognition. The cropped image was used to perform the Siamese network on the image collected by crawling, cropped the part corresponding to the object, labeled, and classified. The backbone network was trained by fine-tuning Mobilenet\_v3\_large[17], a pre-trained model with ImageNet, to fit the current data.

Our Siamese Network is a deep neural network based on MobileNet-v3-large/small[17].

MobileNet[18] is a model that focuses on reducing the number of parameters and operations to fit the mobile phone CPU. MobileNetV1 effectively improved computation efficiency with depthwise separable convolution, and MobileNetV2[19] extended this with a resource-efficient block with inverted residuals and linear bottlenecks. MobileNetV3 proposed an optimal architecture that added squeeze and excitation modules to the bottleneck layer, which achieves high accuracy with similar or lower latency cost than the previous model.

MobileNetV3-Large and MobileNetV3-Small, which are targeted for high and low resource use cases. These models are then adapted and applied to the tasks of object detection and semantic segmentation.

MobileNet has two fully connected layers at the end. Its last fully connected layer is designed regarding the original train dataset class(1000 classes). We do not need to classify images as 1000 classes and need feature representation at lower dimensions. So we remove the last fc layer and replace it with 3 fc layers.

We applied YOLO classification and crop to image preprocessing among the existing network architectures so that the siamese network can achieve the same or higher performance with a lightweight model that requires fewer resources than the existing one. This can be applied to applications that correct the classification error of YOLO after crawling, and additionally, it was shown that the intensity level reflecting the image reality can be adjusted according to the value of the distance threshold applied to the siamese network.

We performed the crop function of YOLO as a pre-processing task of the image input to the siamese network so that the lightweight model can operate as the core network model of the siamese network, which can facilitate the operation of the siamese network on mobile or edge devices.

The user using the application can receive the image that has finally gone through the image matching of the siamese network after crawling the desired image through the object crop function of YOLO. Here, thanks to the object crop function of YOLO, the core model of the siamese network can exhibit sufficient performance even with a lightweight model, and the user can use the application with fewer resources.

    \includegraphics[width=0.5\linewidth]{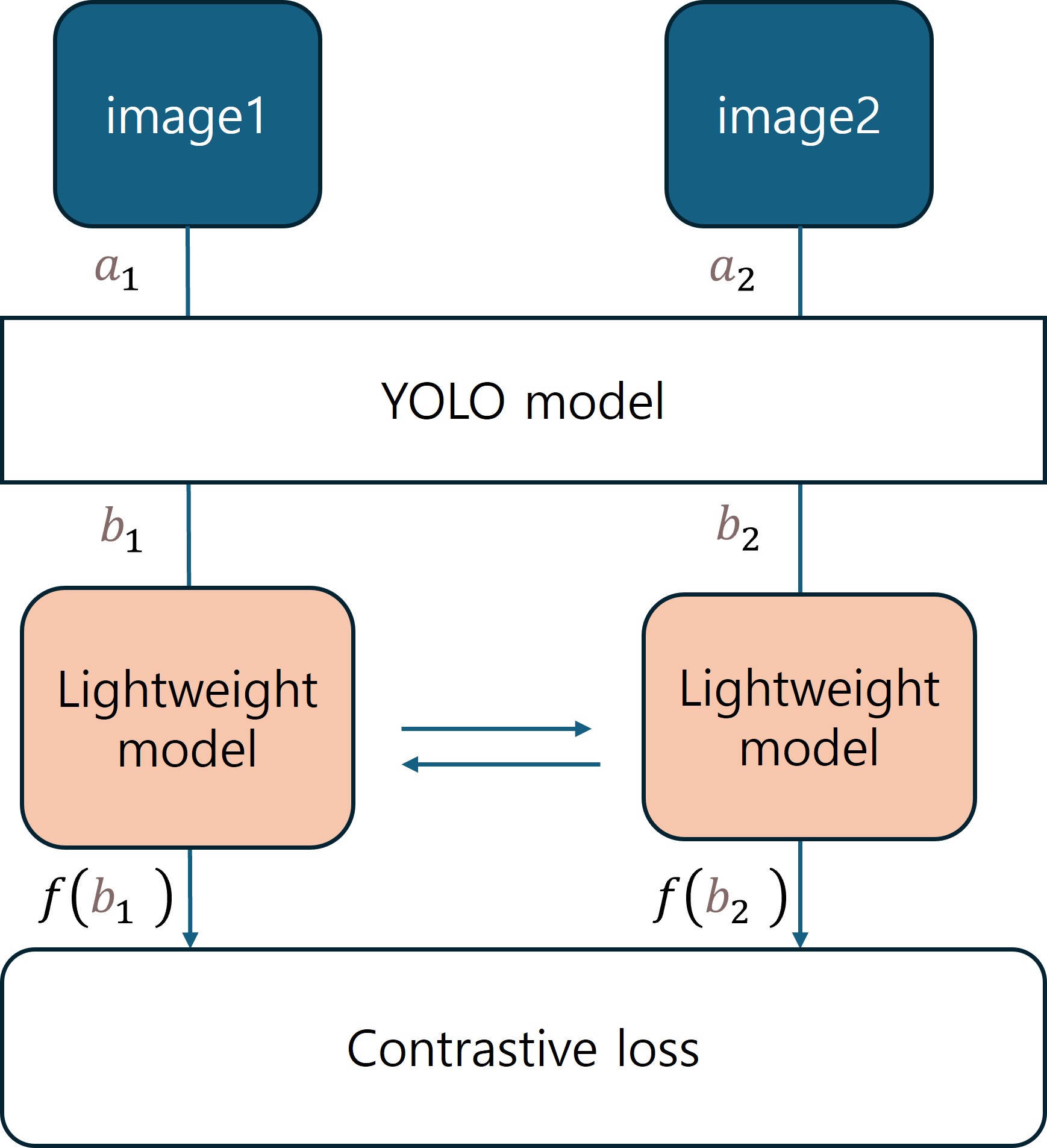}

\textbf{Fig.8. Lightweight Siamese Network Architecture}
 image1, image2: The feature representations of each image, denoted as a1 and a2 are passed to the YOLO model.
 
YOLO model: The YOLO model detects objects from the two input images and performs cropping operations on the objects. In this process, cropped image features denoted as b1 and b2 are generated.

Lightweight model : b1 and b2 become f(b1) and f(b2) feature vectors through low-resource MobileNetV3-small model.

Contrastive loss: Finally, the contrastive loss function is used to learn the differences between the two images. This loss function calculates the similarity/dissimilarity between the two images, helping the Siamese network learn the relationship between the two images. We applied YOLO classification and crop to image preprocessing among existing network architectures, so that the lightweight MobileNetV3-small model, which requires less resources than the existing one, can achieve the same or higher performance than when the siamese network uses MobileNetV3-large without preprocessing. We performed YOLO's crop function as a pre-processing task for the image input to the siamese network, so that the lightweight model can operate as the core network model of the siamese network, which can facilitate the operation of the siamese network on mobile or edge devices.

This can be applied to applications that correct YOLO's classification error after crawling, and additionally, the intensity level reflecting the image reality can be adjusted according to the value of the distance threshold applied to the siamese network.

Users of the application can crawl the desired image, then receive an image that has undergone the YOLO object crop function and the final image matching of the siamese network. Here, thanks to YOLO's object cropping function, the core model of the siamese network can achieve sufficient performance even with a lightweight model, allowing users to use applications with fewer resources.

\section{\textbf{\textbf{\textbf{IV. EXPERIMENTS} } }}

\textbf{A. Image similarity metric}
The f1\_N score, which modifies the F1 score by changing the evaluation criteria of the model, serves to supplement the F1 score. We use the f1\_N score to solve the data imbalance problem. The crawled data is data obtained as a result of a keyword search on Google Images. Therefore, most of the images are composed of positive examples. When positive data overwhelms negative data, the skew rate, which is the ratio of negative to positive data, approaches 0. If the skew rate is too close to 0, the F1 score will be biased toward 1 regardless of the error rate. (Refer to Figure 4, where the error rates of 1\% and 20\% recorded F1 scores of 1 and 0.9, respectively.) This makes it difficult to distinguish the difference in F1 scores according to the error rate of the classifier.

    \includegraphics[width=0.5\linewidth]{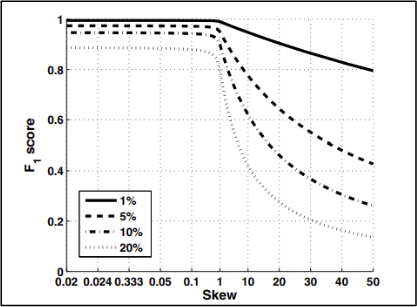}

\textbf{ Fig.9. f1-score difference by simulated classifier’s error rate [20]}

In model labeling, we change the positive and negative (TP -> TN, FP -> FN) to make the skew rate greater than 1. In the f1\_N score metric, precision and recall are changed to negative predictive value (NPV) and specificity, respectively. However, if the skew rate in the f1\_N score is too large, the ratio may be excessively distributed, as shown in Figure 4. Therefore, we use the average\_f1 score, which uses the f1 score and the f1\_N score together to create a balanced metric. 

\[ Skew=\frac{nagative\;examples}{positive\;examples}\]\;\;\;\;\;\;\;\;\;\;\;\;\;(a)

\[Average\_f1\;score=\frac{F1\;score+F1\_N\;score}{2}\]

\[ F1\_N\;score=2\times\frac{NPV\times specificity}{NPV+specificity}\]

\[F1\;socre=2\times\frac{precision\times recall}{precision+recall}\]
\;\;\;\;\;\;\;\;\;\;\;\;\;(b)

\textbf{Fig.10. (a) skew rate  (b) evaluation metrics of Average\_F1 score, mean of F1\_N score and F1 score} 
\textbf{B. Data and data preprocessing}

\textbf{Train dataset description:}
The total number of datasets used for Siamese Network and MobileNet\_V3 training is 26,360 (train: 20,709, validation: 5651). The object part was cut out from the existing VOC data used in YOLO and extracted according to the number of each class. The number of anchor images used for each train and validation is 1, and the rest are other images. At this time, the number of support sets (classes) used for others is 20 in total, including keyword labels and non-keyword labels. If it is a positive image, the Euclidean distance with the anchor image is labeled as 0, and if not, it is labeled as 1. Also, the ratio of images similar to anchors and images different from other images in the other image is approximately 1:19.

\textbf{train dataset preprocessing process:}
Test dataset description: 1) Crawling dataset: 100 random images from the real world that can be in 20 classes corresponding to the voc dataset were crawled from Google Image Search and entered into Shamnet to calculate the Euclidean distance with the anchor.

\textbf{C. Experiment Details}

ELU: It is a function with a form almost similar to ReLU. However, unlike ReLU, it solves the Dying ReLU problem by smoothly bending when it is less than 0.[21]

learning rate: 0.000015

epoch : 30(We selected the best model of all 30 models.)

Learning environment : CPU-AMD Ryzen 5 5600X, GPU-NVIDIA GeForce RTX 3070, REM-16G(Siamese Network, MobileNet\_V3) or CPU-14700F, GPU-NVIDIA GeForce RTX 4070, REM-32G (YOLOv10)

\section{\textbf{\textbf{\textbf{\textbf{V. RESULTS AND DISCUSSION} } } }}

\textbf{(1)YOLO+Siamese Network} 

    \includegraphics[width=0.75\linewidth]{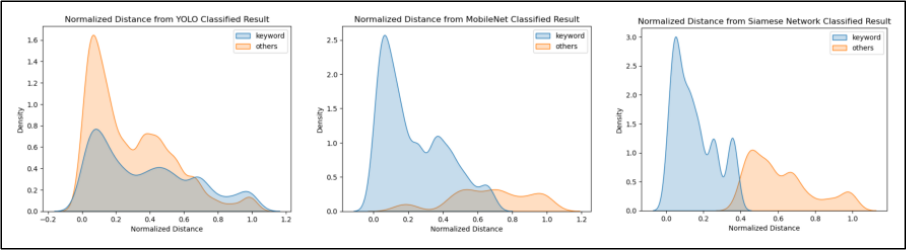}

(a)   \; \;\;\;\;\;\;\;\;\;\;\;\;\;\;\;\;\;\;\;\;\;\;\;\;\;\;\;\;\;\;\;\;\;\;\;\;\;\;\;\;\;\;\;\;(b)       \;\;\;\;            \;\;\;\;\;\;\;\;\;\;\;\;\;\;\;\;\;\;\;\;\;\;\;\;\;\;\;\;\;\;\;\;\;\;\;                 (c)

\textbf{Fig. 11. Distance Density of what is classified as a keyword or other class from (a) YOLO (b) YOLO+mobilenet (c) YOLO+siamese network(mobilenet)}

Using the trained siamese network, we can obtain the Euclidean distance between the anchor image for each class of the VOC dataset and 100 randomly crawled images.

In Figure.3, for 100 crawled test sets, (a) YOLO, (b) MobileNet, and (c) a Siamese network using MobileNet as a backbone show the distance histogram for positive and negative pairs of each model, with a purple distribution when classified as a keyword and an orange distribution when classified as others. (a) YOLO, (b) MobileNet classify as keywords if the classified label is correct, and as others if it is another label. (c) The Siamese network finds the distance threshold that maximizes the f1-score and classifies it as a keyword if it is smaller than this and as others if it is larger.

Here, we can see that the distance cohesion for the keywords classified by the model is higher when MobileNet is applied to the Siamese network than when YOLO and MobileNet are applied to classification. The Euclidean distance between the images classified by keywords and the correct anchor image is gathered, which means that images similar to the anchor image are distributed small and different images are distributed large.

    \includegraphics[width=0.75\linewidth]{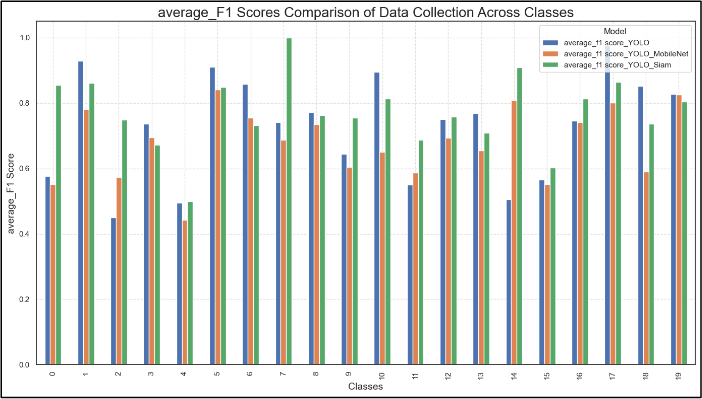}
    
    (a)
    
    \includegraphics[width=0.5\linewidth]{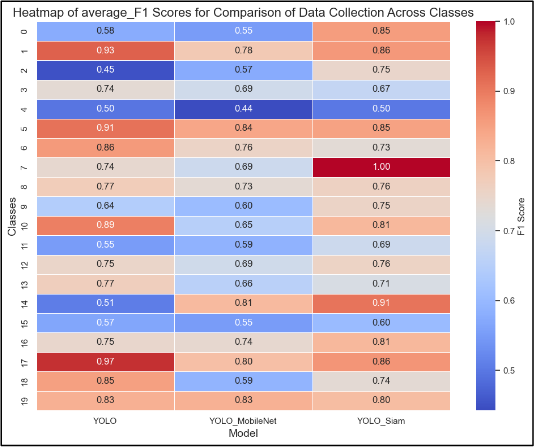}
   
    (b)

\textbf{Fig. 12. Evaluation of average\_f1 score based on different data collection methods on 100 web-crawled test dataset (a) values and (b) heatmap of all 20 classes}

This is an additional model performance evaluation, focusing on various aspects of the classification task.

Fig. 3 compares the performance of three models (YOLO, YOLO+MobileNet, and Siamese Network with YOLO+MobileNet as a backbone). The average f1 score is a balanced index that averages the f1n\_score, which reflects the data imbalance with less negative data and the original f1\_score. (a) is a bar graph comparing the F1 scores for each class for the three models. The F1 scores for each class are shown in blue, YOLO in orange, and Siamese Network in green. From the graph, we can see that Siamese Network recorded the highest F1 scores in many classes. This point emphasizes that the Siamese network performs better in similarity-based tasks between images, showing that similar images are grouped closer together and dissimilar images are more dispersed.

(b) is a heatmap showing the average F1 score for each class, where darker colors indicate higher F1 scores and lighter colors indicate lower scores. The heatmap shows that the combined YOLO and Siamese Network method achieved higher F1 scores in most classes, showing better classification performance than other models. This result suggests that the Siamese Network clustered more homogeneous images in the Euclidean distance-based classification between anchor images and other images after YOLO classification.

It can be seen that Siamese Network classifies similar and dissimilar images well and is more effective than simply reclassifying with YOLO or MobileNet after YOLO.

\textbf{2)Effect of Crop pre-processing and backbone model on Siamese Network performance} 

    \includegraphics[width=0.95\linewidth]{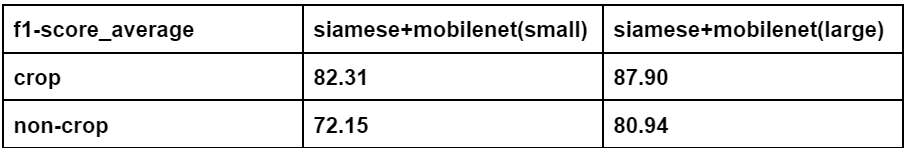}
    
    (a)

    \includegraphics[width=0.75\linewidth]{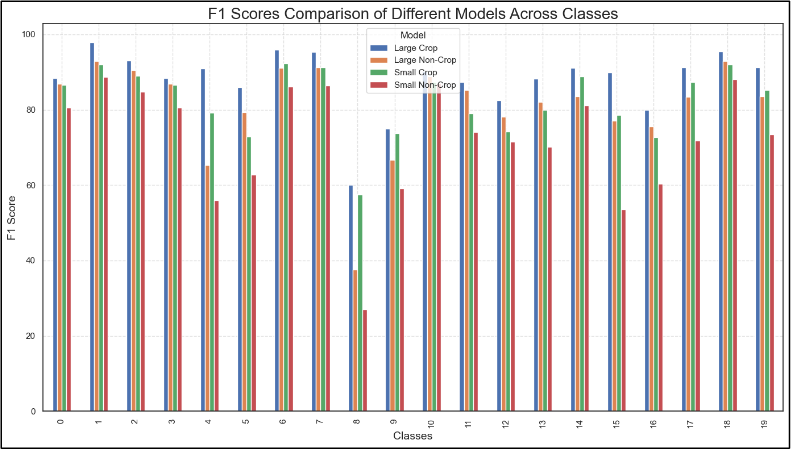}
    
    (b)

    \includegraphics[width=0.5\linewidth]{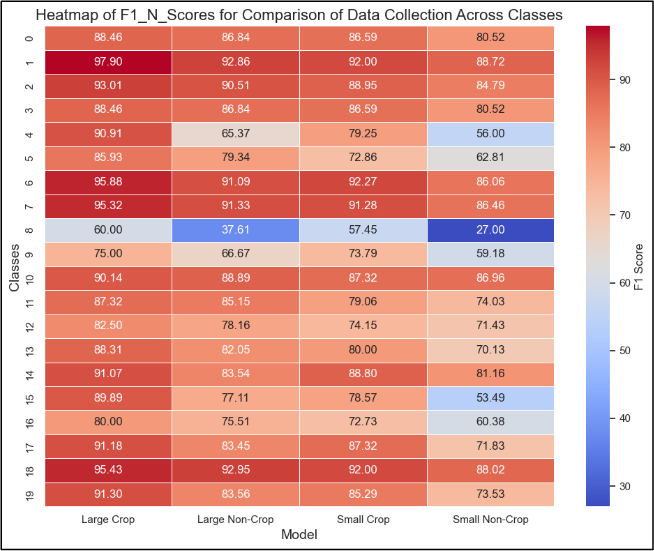}
    
    (c)

\textbf{Fig. 13. Evaluation of Siamese network performance f1-score depending on conditions of Mobilenet\_v3\_small/large and crop/non-crop pre-processing of VOC validation dataset (a) mean value (b) values (c) Heatmap of all 20 classes}

The above Fig. 4. (a) is a table that averages the f1-scores when the backbone of Siamese network is set to Mobilenet\_v3\_small and the training data is cropped or not, and when the backbone of Siamese network is set to Mobilenet\_v3\_large and the training data is cropped or not for all 20 classes of the VOC dataset. (b) is a bar graph for all classes, and (c) is a Heatmap. The train and validation datasets were composed of only those with one object in the photo extracted from the existing VOC dataset. This is because before cropping the object part with YOLO, the Siamese network cannot compare the image with the anchor image if there are multiple objects in the image.

In (a), when the image was pre-cropped to fit the object in our experiment, the Siamese network with the Mobilenet\_v3\_small model as the backbone showed a performance increase of 1.69\% compared to the performance of 80.94 when the non-cropped image was passed through the Siamese network with the Mobilenet\_v3\_large model as the backbone, which was 82.31. This shows that when the image is cropped and passed through the Siamese network based on the classification model, the performance is similar to that of Mobilenet\_v3\_large, or even higher. Also, when looking at (b) and (c), we can see that the performance difference between the Siamese network model that trained on cropped data in some classes, such as class 4 and class 8, and the Siamese network model that trained on non-cropped data using the Mobilenet\_v3\_small model as a backbone is significant. In addition, whether Mobilenet\_v3\_small or Mobilenet\_v3\_large is used as a backbone, the performance is higher when cropped images are inserted than when non-cropped images are inserted. In particular, when Mobilenet\_v3\_small is used, the performance increases by 14.08\%, while when Mobilenet\_v3\_large is used, the performance increases by only 8.5\%. In comparison, the Siamese network using Mobilenet\_v3\_small as a backbone has 1,030,224 parameters, and the Siamese network using Mobilenet\_v3\_large as a backbone has 3,130,464 parameters, which is about 3 times different. That is, we can see that the performance improvement is more noticeable in lightweight models.

\section{\textbf{\textbf{\textbf{\textbf{VI. CONCLUSION} } } }}
The authors of this paper designed a web crawling framework using YOLO for data crawling. In addition, the authors found that using a Siamese network can improve performance compared to using only YOLO or YOLO + MobileNet. In addition, using YOLO, it is possible to obtain various types of data that contain pictures of specific objects or specific objects. As a result, this framework can obtain a lot of high-quality data that suits the user's purpose in a short period of time compared to existing web crawling methods. In addition, the authors found through experiments that when cropped data from a Siamese network is used, it can show performance equivalent to or higher than that of a large model trained on relatively small-scale, uncropped original data. If this fact is applied, we can expect performance improvements in various fields that are currently using the Siamese network.

\section{\textbf{\textbf{\textbf{\textbf{VII. REFERENCE} } } }}
[1]Song, H., Kim, M., Park, D., Shin, Y., \& Lee, J. G. (2022). Learning from noisy labels with deep neural networks: A survey. \textit{IEEE transactions on neural networks and learning systems}, \textit{34}(11), 8135-8153.

[2]Goodfellow, I. J., Shlens, J., \& Szegedy, C. (2014). Explaining and harnessing adversarial examples. \textit{arXiv preprint arXiv:1412.6572}.

[3]Wang, X., Yang, T., Chen, G., \& Lin, Y. (2014). Object-centric sampling for fine-grained image classification. \textit{arXiv preprint arXiv:1412.3161}.

[4]Kang, W., Mun, J., Lee, S., \& Roh, B. (2023). Noise-aware learning from web-crawled image-text data for image captioning. In \textit{Proceedings of the IEEE/CVF International Conference on Computer Vision} (pp. 2942-2952).

[5]Gregory Koch, Richard Zemel, Ruslan Salakhutdinov. Siamese Neural Networks for One-shot Image Recognition https://www.cs.cmu.edu/\~rsalakhu/papers/oneshot1.pdf

[6]Kale, Y. V., Shetty, A. U., Patil, Y. A., Patil, R. A., \& Medhane, D. V. (2021, December). Object detection and face recognition using yolo and inception model. In \textit{International Conference on Advanced Network Technologies and Intelligent Computing} (pp. 274-287). Cham: Springer International Publishing.

[7]Bakhshayeshi, I., Erfani, E., Taghikhah, F. R., Elbourn, S., Beheshti, A., \& Asadnia, M. (2023). An Intelligence Cattle Reidentification System Over Transport by Siamese Neural Networks and YOLO. \textit{IEEE Internet of Things Journal}, \textit{11}(2), 2351-2363.

[8]Melekhov, I., Kannala, J., \& Rahtu, E. (2016, December). Siamese network features for image matching. In \textit{2016 23rd international conference on pattern recognition (ICPR)} (pp. 378-383). IEEE.

[9]Faheem, M. (2012, April). Intelligent crawling of Web applications for Web archiving. In \textit{Proceedings of the 21st International Conference on World Wide Web} (pp. 127-132).

[10]Manku, G. S., Jain, A., \& Das Sarma, A. (2007, May). Detecting near-duplicates for web crawling. In \textit{Proceedings of the 16th international conference on World Wide Web} (pp. 141-150)

[11]Tao, Q., Yang, H., \& Cai, J. (2018). Exploiting web images for weakly supervised object detection. \textit{IEEE Transactions on Multimedia}, \textit{21}(5), 1135-1146..

[12]Hwang, K. H., Lee, M. J., \& Ha, Y. G. (2020, February). A befitting image data crawling and annotating system with cnn based transfer learning. In \textit{2020 IEEE International Conference on Big Data and Smart Computing (BigComp)} (pp. 165-168). IEEE.

[13]Hwang, J., Kim, J., Chi, S., \& Seo, J. (2022). Development of training image database using web crawling for vision-based site monitoring. \textit{Automation in Construction}, \textit{135}, 104141.

[14]Ao Wang, Hui Chen, Lihao Liu, Kai Chen, Zijia Lin, Jungong Han, Guiguang Ding. YOLOv10: Real-Time End-to-End Object Detection \href{https://arxiv.org/abs/2405.14458}{https://arxiv.org/abs/2405.14458}

[15]\href{http://host.robots.ox.ac.uk/pascal/VOC/}{http://host.robots.ox.ac.uk/pascal/VOC/}

[16]Hadsell, R., Chopra, S., \& LeCun, Y. (2006, June). Dimensionality reduction by learning an invariant mapping. In \textit{2006 IEEE computer society conference on computer vision and pattern recognition (CVPR'06)} (Vol. 2, pp. 1735-1742). IEEE

[17]Howard, A., Sandler, M., Chu, G., Chen, L. C., Chen, B., Tan, M., ... \& Adam, H. (2019). Searching for mobilenetv3. In \textit{Proceedings of the IEEE/CVF international conference on computer vision} (pp. 1314-1324).

[18]Andrew G. Howard, Menglong Zhu, Bo Chen, Dmitry Kalenichenko, Weijun Wang, Tobias Weyand, Marco Andreetto, and Hartwig Adam. Mobilenets: Efficient convolutional neural networks for mobile vision applications. CoRR, abs/1704.04861, 2017.

[19] Mark Sandler, Andrew G. Howard, Menglong Zhu, Andrey Zhmoginov, and Liang-Chieh Chen. Mobilenetv2: Inverted residuals and linear bottlenecks. mobile networks for classification, detection and segmentation. CoRR, abs/1801.04381, 2018

[20]Jeni, L. A., Cohn, J. F., \& De La Torre, F. (2013, September). Facing imbalanced data--recommendations for the use of performance metrics. In \textit{2013 Humaine association conference on affective computing and intelligent interaction} (pp. 245-251). IEEE.

[21] Djork-Arné Clevert, Thomas Unterthiner, Sepp Hochreiter(2015). Fast and Accurate Deep Network Learning by Exponential Linear Units (ELUs). arXiv:1511.07289

\end{document}